\documentclass[letterpaper, 10 pt, conference, twoside]{ieeeconf}
\usepackage{times}
\usepackage[pdftex]{graphicx}
\usepackage{subfig}
\usepackage{amsmath,amssymb,amsopn,amstext,amsfonts}
\usepackage{cancel}
\usepackage[space]{cite}
\usepackage{pdfsync}
\usepackage{balance}
\usepackage{color}
\usepackage{mathtools}
\usepackage{algorithm}
\usepackage{algorithmic}
\usepackage{bm}

\usepackage{diagbox}
\usepackage{float}
\usepackage{epstopdf}
\usepackage{url}
\usepackage{booktabs}
\usepackage[linkcolor=black,citecolor=black,urlcolor=black,colorlinks=true]{hyperref}
\usepackage{verbatim}
\usepackage{subfig}
\usepackage{bm}
\usepackage{afterpage} 
\makeatletter
\def\tagform@#1{\maketag@@@{\normalsize(#1)\@@italiccorr}}
\makeatother

\graphicspath{{./Figures/}}
\DeclareGraphicsExtensions{.pdf,.png,.jpg,.eps,.svg}
\IEEEoverridecommandlockouts
\vspace{1.4cm}
\title{DOB-based Wind Estimation of A UAV Using Its Onboard Sensor}
\author{Haowen Yu, Xianqi Liang, and Ximin Lyu %
	\thanks{Project supported by the National Natural Science Foundation of China (Grant No.62303495) and Shenzhen Excellent Scientific and Technological Innovation Talent Training Program (Grant No. RCBS20221O08093104017) }
	\thanks{All authors are with the School of Intelligent Systems Engineering, Sun Yat-sen University, Guangzhou, China. (\textit{Corresponding author:} Ximin Lyu)}%
     \thanks{E-mail: {\tt\small lvxm6@mail.sysu.edu.cn}}
}
\begin{document}
\maketitle

\begin{abstract}
Unmanned Aerial Vehicles (UAVs) play a crucial role in meteorological research, particularly in environmental wind field measurements. However, several challenges exist in current wind measurement methods using UAVs that need to be addressed. Firstly, the accuracy of measurement is low, and the measurement range is limited. Secondly, the algorithms employed lack robustness and adaptability across different UAV platforms. Thirdly, there are limited approaches available for wind estimation during dynamic flight. Finally, while horizontal plane measurements are feasible, vertical direction estimation is often missing. 
To tackle these challenges, we present and implement a comprehensive wind estimation algorithm. Our algorithm offers several key features, including the capability to estimate the 3-D wind vector, enabling wind estimation even during dynamic flight of the UAV. Furthermore, our algorithm exhibits adaptability across various UAV platforms. Experimental results in the wind tunnel validate the effectiveness of our algorithm, showcasing improvements such as wind speed accuracy of $0.11$ m/s and wind direction errors of less than $2.8^\circ$. Additionally, our approach extends the measurement range to $10$ m/s.
\end{abstract}

\section{INTRODUCTION}
In recent years, there has been significant interest in obtaining accurate wind vectors (wind speed and wind direction) due to their importance in predicting environmental changes\cite{patrikar2020windIrosPFGFRCFD}. The ability to measure wind vectors in real time and at multiple locations is crucial for several applications, such as predicting rainfall trends and typhoon log-in dates. Currently, wind measurements are primarily conducted using anemometers and weather balloons. However, anemometers can only provide fixed-point measurements and lack the necessary flexibility. Meanwhile, weather balloons are cumbersome to deploy and expensive since they cannot be reused.

To address these challenges, the utilization of Unmanned Aerial Vehicles (UAVs) for wind vector estimation has gained attention \cite{cho2011windHybridSensorwind}. However, several limitations remain, including low measurement accuracy, a limited measurement range, a lack of robustness and adaptability across different UAV platforms, a scarcity of approaches for in-flight wind estimation, and the absence of vertical direction estimation. 

\begin{figure}[t]
	\centering
	\includegraphics[width=0.88\columnwidth]{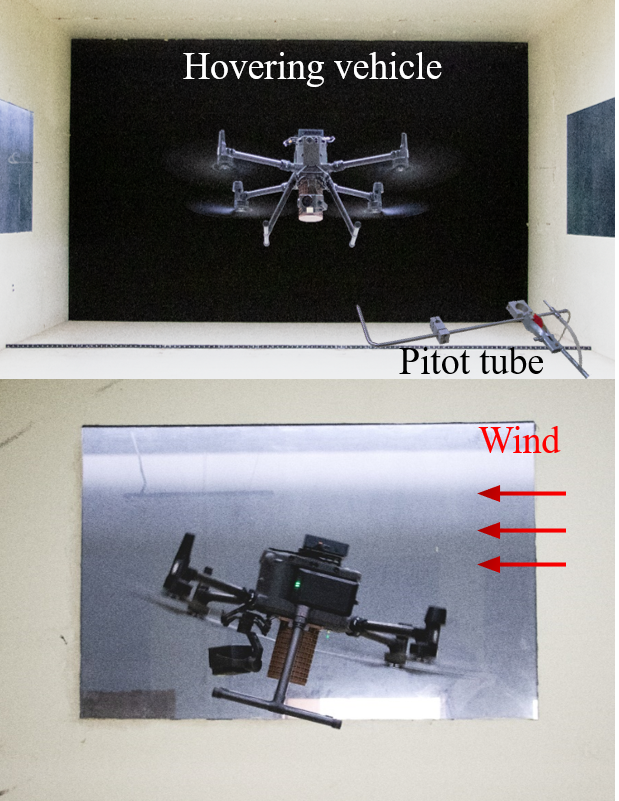}
	\caption{Wind tunnel experiments with a given wind speed, the UAV remains hovering to collect data.
    Video available at: \url{https://youtu.be/QpbzR2NJULg}
		\label{fig:first_page_show}}
	\vspace{-0.49cm}
\end{figure}

Motivated by these issues, we propose a robust and high accurate wind estimation method. Our approach employs a disturbance observer (DOB) to estimate the external force vector exerted by the airflow blowing through the UAV, inspired by \cite{lyu2018dobHinfinity}. By mapping the external force vector to the relative air vector, we obtain a comprehensive 3-D force-air model that can estimate the relative air vector online. Finally, by incorporating the ground speed of the vehicle and the estimated relative air vector, we can derive the wind vector through the synthesis of wind triangles.
Experimental results in the wind tunnel demonstrate that our model has an error of less than $0.11$ m/s for the estimated wind speed and less than $2.8^\circ$ for the wind direction. The robust one-to-one mapping of our three-dimensional (3-D) force-air model ensures resilience to perturbations, enhancing the reliability of our estimation method.
Our proposed method requires only basic inputs such as ground velocity, acceleration, attitude, motor rotation speed, and vehicle mass, which can be easily obtained from a UAV platform. This feature ensures the adaptability of our method to different platforms.
Moreover, leveraging the self-convergence capability of the utilized DOB method\cite{chen2000nonlinearChenwenhua}, our approach enables dynamic calculation of external forces, facilitating real-time wind estimation during dynamic flight operations. To be specific, our research makes the following key contributions:
\begin{enumerate}
\item \textbf{Innovative wind estimation framework}: Our framework consists of a front-end module and a back-end module. The front-end module employs a DOB method to accurately estimate the external force vector. On the other hand, the back-end module utilizes the estimated force vector to perform online estimation of the relative air vector and computes the wind vector by synthesizing wind triangles \cite{NEUMANN2015tiltearliest}. The decoupling of the front-end and back-end modules enables separate optimization for each module using different methods, ensuring system stability and facilitating maintenance.
    
\item \textbf{Improved accuracy and expanded measurement range}: In our experiments conducted with the presented platform, we tested wind speeds up to $10 $ m/s. The mean velocity error was measured to be $0.11 $ m/s, which is less than $2\%$ of the true value. The angle error was found to be less than $2.8^\circ$ in the wind tunnel experiment.

\item \textbf{Estimation of a 3-D wind vector}: The DOB-based estimator allows for the estimation of a 3-D force vector, which in turn enables the estimation of a 3-D wind vector. This capability significantly broadens the scope of wind estimation and provides a more comprehensive understanding of wind patterns for meteorological research purposes \cite{baker2014lidar3DwindMeasurement}.

\item \textbf{Simultaneous estimation during dynamic flight}:
The DOB-based estimator has the ability to estimate the force vector online while the vehicle is in dynamic flight, thereby enabling the accurate estimation of the wind vector in any scenario. This unique feature enhances the efficiency and effectiveness of wind vector estimation, allowing for real-time and continuous monitoring of wind.
\end{enumerate}

The remainder of this paper is organized as follows: Sec.~\ref{sec:related_works} provides an overview of relevant works and methods in the field of wind vector estimation. Sec.~\ref{sec:principle_of_wind_est} presents the principle of wind estimation. Sec.~\ref{sec:dob_design} introduces the proposed DOB design method. The force-air model is established in Sec.~\ref{sec:Force_wind_relationship}. Experimental results are presented in Sec.~\ref{sec:Experiment}. Finally, Sec.~\ref{sec:conclusion} concludes the paper and introduces future work.

\section{RELATED WORK}
\label{sec:related_works}

In the early stages of wind speed and direction measurements, hardware attachments to fixed-wing vehicles, such as pitot tubes \cite{cho2011windfixedwingPitotTube} and custom sensor measurement systems \cite{makaveev2023vtolsensorwind}, were commonly used. However, these sensor systems designed exclusively for fixed-wing configurations proved unsuitable for multi-rotor UAVs due to their limited payload space and the effect of downwash airflow from the rotors.
The dominant approach for estimating wind speed and angle involves judging them solely from the attitude of the UAV when subjected to wind, as highlighted in \cite{NEUMANN2015tiltearliest,palomaki2017tiltwindHighReference}. Rotor UAVs have gained popularity in the field of wind estimation after optimization by K. Meier \textit{et al.} \cite{meier2022windtiltnewAtmosphere} and M. Simma \textit{et al.} \cite{simma2020iris}. However, tilt-based methods have limitations. They can only measure wind speed and direction in the horizontal plane and require the UAV to be in a steady state at all times. 
To address the limitations mentioned above, G. Hattenberger \textit{et al.} \cite{hattenberger2022KFestimating} proposes an estimation method based on Kalman Filter (KF). However, it is crucial to acknowledge that the optimality of the KF heavily depends on the accuracy of the model utilized in the prediction step. If the model is flawed, it can negatively impact the performance of the KF. Additionally, the mathematical models of wind used in these works are often simplistic, limiting the achievable accuracy of the estimation.
To overcome the wind estimation accuracy problem, wind vectors are measured using system identification based on adequate modeling of wind disturbances \cite{gonzalez2020SystemIdentify, loubimov2020SystemIdentify2}. This approach allows for high estimation accuracy. However, excessively complex wind models do not necessarily guarantee robustness.
In recent years, neural network based techniques (NN-based) have been employed to reduce the complexity of wind modeling \cite{ zimmerman2022windMachineLearning1,crowe2020earlyMachineLearning}. By directly predicting wind speed and direction from UAV sensor data, these techniques eliminate the need for explicit wind models. However, the parameter sensitivity of neural networks can lead to large prediction errors when even slight changes occur in the unknown parameters. Consequently, they may exhibit poor adaptability to different models.

\begin{figure}[t]
	\vspace{0.0cm}
	\centering
	\includegraphics[width=0.9\columnwidth]{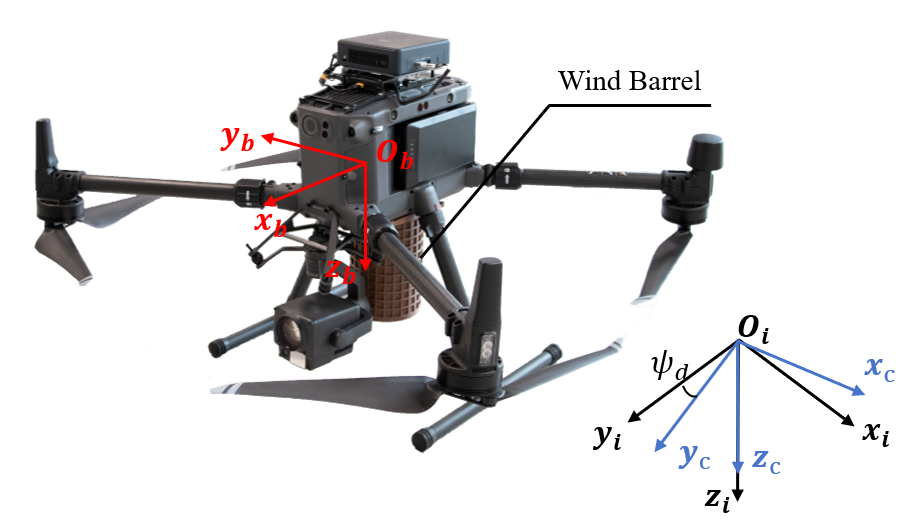}
	\caption{The UAV platform and coordinate system. A wind barrel is installed under the UAV to increase the wind estimation accuracy.
		\label{fig:m300_combine_coord}}
	\vspace{-1.38cm}
\end{figure}

To overcome the limitations of current methods, we introduce a novel DOB-based approach for wind field estimation. The proposed method offers several key advantages \cite{di2022ladcIcraGuo3Disturbance}, including robustness, high accuracy, and the ability to estimate the 3-D wind vector even during the dynamic flight of a UAV. Furthermore, our approach is designed to be easily adaptable to different UAV platforms, ensuring its versatility and applicability across various systems.

\begin{figure*}[t]
	\vspace{0.0cm}
	\centering
	\includegraphics[width=1.9\columnwidth]{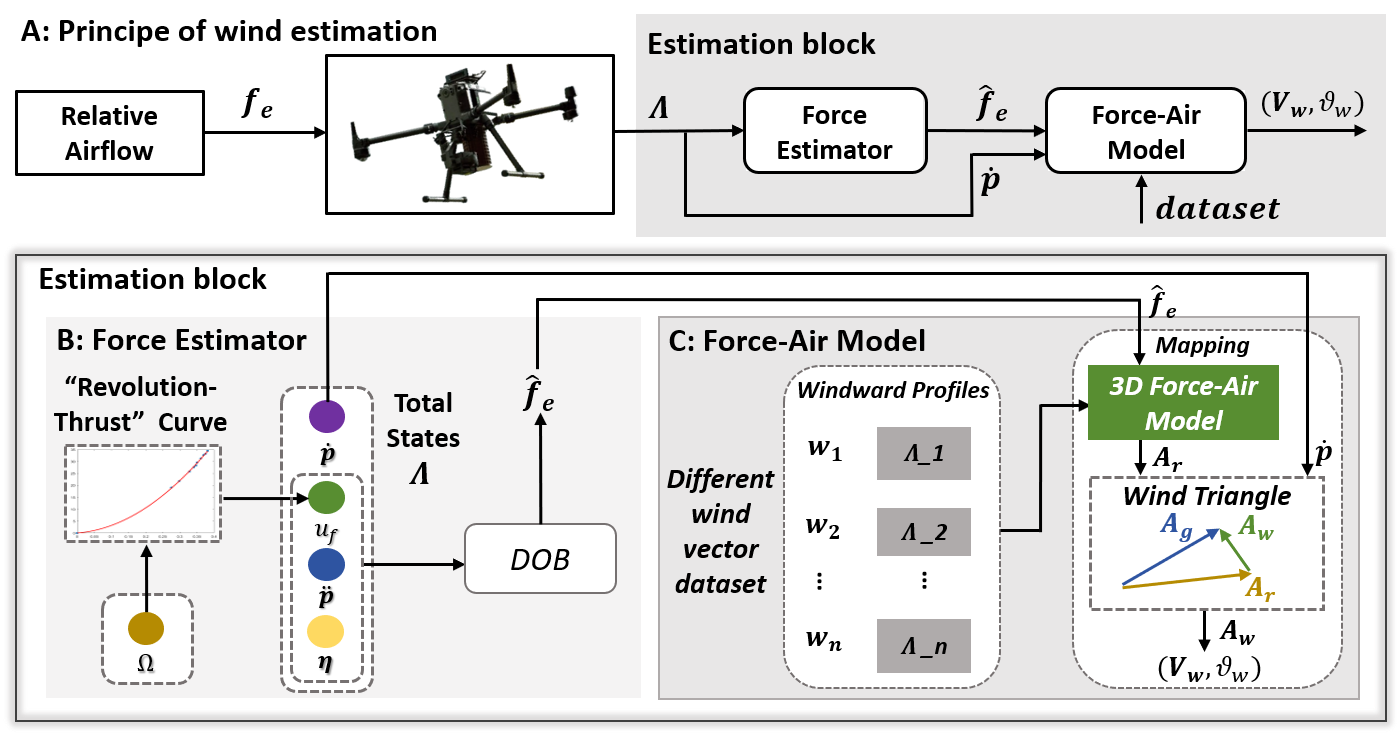}
    
        \caption{Principe of wind estimation: \textbf{(A)} The relative air flow generates a total external force $\bm{f_e}$, which leads to new states $\bm{\Lambda}$. The force estimator uses the real-time states $\bm{\Lambda}$ to acquire the estimated external force $\hat{\bm{f_e}}$. The air-wind model utilizes the ground vector $\dot{p}$ and $\hat{\bm{f_e}}$ to estimate the wind speed $\bm{V_w}$ and wind direction $\vartheta$.   
        \textbf{(B)} Front-end module: The force estimator acquires and processes the total states $\bm{\Lambda}$ to obtain the external force $\hat{\bm{f_e}}$.  
        \textbf{(C)} Back-end module: Build a 3-D force-air model using a pre-calibrated force-air vector dataset. By incorporating the known ground vector $\dot{p}$ and applying the principles of the wind triangle, we can ultimately determine the wind vector and extract the wind speed and direction.
		}
    \label{fig:system_design_frame}
	\vspace{-0.3cm}
\end{figure*}

\section{PRINCIPLE OF WIND ESTIMATION}
\label{sec:principle_of_wind_est}

The coordinate frames of our UAV platform are shown in Fig. \ref{fig:m300_combine_coord}: the body frame $x_b$, $y_b$, $z_b$, the inertial frame $x_i$, $y_i$, $z_i$, and the intermediate frame $x_c$, $y_c$, $z_c$. The origin $o_b$ of the body frame is set to coincide with the vehicle's center of gravity. The intermediate frame is defined by rotating along $z_i$ by $\psi_d$. $\psi_d$ is the desired yaw angle when the vehicle is hovering. 
\begin{figure}[t]
	\vspace{0.2cm}
	\centering
	\includegraphics[width=0.85\columnwidth]{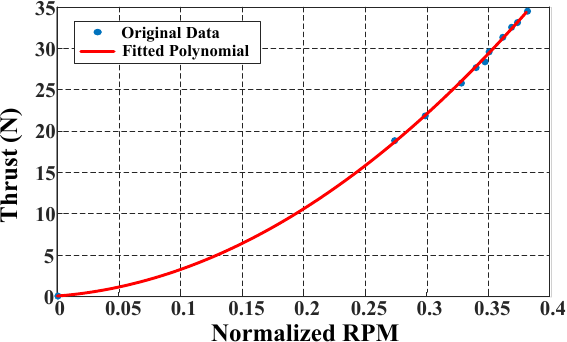}
 \caption{'Revolution-Thrust' curve for each motor.
		\label{fig:speed_thrust_curve}}
	\vspace{-0.15cm}
\end{figure}
The equation of dynamics of the quadrotor can be written in \eqref{equ:dynamic1} as follows \cite{emran2018dynamicmodellagelangri}:
\begin{equation}
\label{equ:dynamic1}
\begin{aligned}
    &m\bm{\ddot{p}}=-u_f\bm{R(\bm{\eta})}\bm{e_3}+m{g_0}\bm{e_3}+\bm{\bm{f_e}}\\
    &\bm{J}\bm{\ddot{\eta}}=-\bm{\mathbb{C}(\eta,{\dot{\eta}}){\dot{\eta}}}+\bm{u_{\tau}}+\bm{{\tau}_e}\\
\end{aligned}
\end{equation}
where $m$ is mass; $g_0$ is gravitational constant; ${\bm{J}}\in {{\mathbb{R}}^{3\times 3}}$ is inertia matrix in the body frame; $\bm{p} = [p_x,p_y,p_z]^T$ is the position of the quadrotor in the inertial frame; $\bm{\eta} = [\phi,\theta,\psi]^T$ represents roll, pitch, and yaw angle, respectively; $\bm{R(\eta)}$ is the rotation matrix from body frame to inertial frame; $\bm{e_3} = [0,0,1]^T$; $\bm{\mathbb{C}(\eta,{\dot{\eta}})}$ is the coriolis matrix; $\bm{{{f}_{e}}}= [f_{ex},f_{ey},f_{ez}]^T$ is total external force vector in inertial frame; ${\bm{\tau}}_{\bm{e}}\in {{\mathbb{R}}^{3}}$ is total external torque vector in the body frame; ${{u}_{f}} \in {{\mathbb{R}}^{1}}$ and $\bm{{{u}_{\bm{\tau}}}}\in {{\mathbb{R}}^{3}}$ are respectively total thrust and torque which is produced by the thrust of the four rotors. Fig.~\ref{fig:speed_thrust_curve}, illustrates the relationship between the thrust and revolutions per minute (RPM) of each rotor. This relationship can be approximated using a quadratic polynomial\cite{deters2014reynoldsLeinuoshu2}. In our specific scenario, the motor revolution-thrust function can be expressed as:
\begin{equation}
\label{equ:thrust_curve}
\begin{aligned}
    {u_f} = \sum_{i=1}^4 (207\Omega_i^2+11.34\Omega_i+0.01315)
\end{aligned}
\end{equation}
where $\Omega_i$ is the normalized RPM which represents the current RPM divided by the max RPM.

As shown in Fig.~\ref{fig:system_design_frame} (A), our proposed wind estimation method is based on a straightforward concept. When the vehicle is hovering or flying in the sky, the relative airflow around the vehicle generates a total external force $\bm{f_e}$. By estimating this external force and establishing the relationship between the external force vector and the relative air vector, we can map the relative air vector based on $\bm{f_e}$. With knowledge of the relative air vector and the ground velocity, we can determine the wind speed $\bm{V_w}$ and wind direction $\vartheta_w$. Importantly, our wind estimator allows real-time estimation of $\bm{f_e}$, making it suitable for both hovering and dynamic flight scenarios. 

Fig.~\ref{fig:system_design_frame} (B) illustrates the process of force estimation. We employ a nonlinear disturbance observer (DOB) to estimate the external force. The DOB utilizes the total thrust $u_f$, acceleration $\bm{\ddot{p}}$, and attitude $\bm{\eta}$ of the vehicle to calculate the $\bm{f_e}$. The detailed design and implementation of the DOB will be discussed in Sec.~\ref{sec:dob_design}.

Fig. \ref{fig:system_design_frame} (C) provides a detailed explanation of the process for obtaining wind information. We conduct a wind tunnel test to collect pre-calibrated relative air vectors $\bm{A_r}$ and total states $\bm{\Lambda} = [\bm{\dot{p}}, \bm{\ddot{p}}, \bm{\eta}, \Omega]$, which are used to establish a 3-D force-air model. Given the $\bm{f_e}$, we utilize the 3-D force-air model to map the relative air vector $\bm{A_r}$. By considering the relative air vector $\bm{A_r}$ and the ground vector $\bm{A_g} = \bm{\dot{p}}$, and employing the principles of wind triangle synthesis (i.e., $\bm{A_r}=\bm{A_w}+\bm{A_g} $), we can determine the wind vector $\bm{A_w}$ and extract both the wind speed $\bm{V_w}$ and wind direction $\vartheta_w$. For detailed calculations, please refer to Sec.~\ref{sec:Force_wind_relationship}.

\section{DOB DESIGN}
\label{sec:dob_design}
Due to the fact that the same wind generates different external forces in various directions of the vehicle, it results in distinct convergence rates. These different convergence rates can lead to errors in wind estimation. To ensure consistent convergence rates for the estimated force along the XYZ axes, we propose our DOB method based on the work of B. Y{\"u}ksel \textit{et al.}~\cite{yuksel2014nonlinearFourwrench}. We extend their approach and provide a proof of stability for our modified DOB.
We write \eqref{equ:dynamic1} in a compact form:
\begin{equation}
\label{equ:wrench_observer_lagelangri}
\begin{aligned}
    \bm{d_e}=\bm{B}\bm{\ddot{q}}+\bm{D}(\bm{q},\bm{\dot{q}})\bm{\dot{q}}+\bm{g}-\bm{G(\bm{q})}\bm{u}
\end{aligned}
\end{equation}
where
\begin{equation}
\label{equ:dynamic3}
\begin{aligned}
    \bm{B} &= \left[ \begin{matrix}
	m\bm{I} & \bm{0}  \\
	\bm{0} & \bm{J}  \\
	\end{matrix} \right] \bm{D}(\bm{q},\bm{\dot{q}}) = \left[ \begin{matrix}
	\bm{0} & \bm{0}  \\
	\bm{0} & {\bm{\mathbb{C}(\eta,{\dot{\eta}})}}  \\
	\end{matrix} \right] \\
    \bm{g} &= \left[ \begin{matrix}
	-m{g_0}\bm{e_3}  \\
	\bm{0}   \\
	\end{matrix} \right] \bm{G(\bm{q})} = \left[ \begin{matrix}
	-\bm{R(\bm{\eta})}\bm{e_3} & \bm{0} \\
	\bm{0} & \bm{I} \\
	\end{matrix} \right]
\end{aligned}
\end{equation}
where $\bm{d_e} = [\bm{{f_e}}^T, \bm{{\bm{\tau}_e}}^T]^T$ is an external wrench that acts on the body of the UAVs; ${\bm{I}}\in {{\mathbb{R}}^{3\times 3}}$ is identity matrix; $\bm{q} = [\bm{{p}}^T, \bm{{\eta}}^T]^T$ is the quadrotor's posture; $\bm{u} = [{u_f}, {\bm{u_\bm{\tau}}}^T]^T \in {{\mathbb{R}}^{4\times 1}}$. We propose the following DOB-based on \cite{chen2000nonlinearChenwenhua}:
\begin{equation}
\label{equ:we_derivative_from_chenwenhua}
\begin{aligned}
    \bm{\dot{\widehat{d_e}}}=\bm{L(q,{\dot{q}})(d_e-\widehat{d_e})}
\end{aligned}
\end{equation}  
where $\bm{\widehat{d_e}}$ is the estimated wrench, $\bm{d_e}$ is the true value introduced by \eqref{equ:wrench_observer_lagelangri}. In general, there is no prior information about the derivative of the disturbance, it is reasonable to suppose that:
\begin{equation}
\begin{aligned}
\label{equ:we_equals_zero0}
    \bm{\dot{d_e}}=0
\end{aligned}
\end{equation}  
we define the observation error as the difference between the true and observed values:
\begin{equation}
\begin{aligned}
\label{equ:observer_error}
    \bm{e={d_e}-\widehat{d_e}}
\end{aligned}
\end{equation}  
combining \eqref{equ:we_derivative_from_chenwenhua} and \eqref{equ:observer_error}, we can now calculate:
\begin{equation}
\begin{aligned}
\label{equ:doteplusLe_equals_dotwe}
   \bm{ \dot{e}= {L(q,{\dot{q}})}{\widehat{d_e}} - {L(q,{\dot{q}})}{d_e} = -{L(q,{\dot{q}})}e}
\end{aligned}
\end{equation}

Therefore, the ${\bm{L(q,{\dot{q}})}}$ directly impacts the convergence of error dynamics. In order to achieve different estimation effects for $\bm{f_e}$ on XYZ axes respectively, inspired by \cite{yuksel2014nonlinearFourwrench}, we design the following observer: 
\begin{equation}
\begin{aligned}
\label{equ:Lqq_define}
    \bm{ L(q,{\dot{q}})={K_I}{B^{-1}} }
\end{aligned}
\end{equation}
where $\bm{K_I} \in {{\mathbb{R}}^{3\times 3}}$ is the observation parameter matrix, a diagonal matrix where the elements on the main diagonal can be different parameters.

\textit{Proposition 1: } Consider the wrench estimator from \eqref{equ:we_derivative_from_chenwenhua}. If $\bm{L(q,{\dot{q}})}$ is defined as \eqref{equ:Lqq_define}, then $\bm{\widehat{d_e}} \rightarrow \bm{d_e}$. 

\textit{Proof: } We will demonstrate the convergence of the error in equation \eqref{equ:observer_error} by establishing the asymptotic stability of \eqref{equ:doteplusLe_equals_dotwe}:
\begin{equation}
\begin{aligned}
    \bm{ V(e,q)={e^T}\bm{B}e }
\end{aligned}
\end{equation}
be a positive definite candidate Lyapunov function. We can write:
\begin{equation}
\label{equ:Liyapunuofu}
\begin{aligned}
    \frac{d\bm{V(e,q)}}{dt}= -2 \bm {{e^T}B{K_I}{B^{-1}e}+{e^T}\dot{B}e }\\
    = \bm{{e^T}} \bm{(} -2 \bm {B{K_I}{B^{-1}}+\dot{B})e}
\end{aligned}
\end{equation}
where $\bm{B}$ is symmetric, $m\bm{I}$ and $\bm{J}$ are both constant values. Therefore, we can compute that:
\begin{equation}
\label{equ:Bq_dot}
\begin{aligned}
    \bm{\dot{B}} = \left[ \begin{matrix}
	\bm{0} & \bm{0}  \\
	\bm{0} & \bm{0}  \end{matrix} \right]
\end{aligned}
\end{equation}

We can infer \eqref{equ:Liyapunuofu} that:
\begin{equation}
\begin{aligned}
\label{equ:KI_condition}
    \frac{d\bm{V(e,q)}}{dt}=
    \bm{{e^T}}\bm{(}-2 \bm {B{K_I}{B^{-1}})e}
\end{aligned}
\end{equation}
since $\bm{V(e,q)}$ is positive definite, and considering that \eqref{equ:KI_condition} can be negative definite while $\bm{K_I} \in {{\mathbb{R}}^{+}}$, the Lyapunov condition holds. Consequently, $\bm{K_I}$ is positive definite can guarantee the convergence of $\bm{e}(t)$ to $0$, proving the statement. Based on this, we substitute \eqref{equ:Lqq_define} into \eqref{equ:we_derivative_from_chenwenhua} yields:
\begin{equation}
\label{equ:observer_matrix_equation}
\begin{aligned}
    \bm{\dot{\widehat{d_e}}}=  -\bm{K_I} \bm{B^{-1}}\bm{\hat{d_e}}+
    \bm{K_I} \bm{B^{-1}}(\bm{B}\bm{\ddot{q}}+\bm{D}\bm{\dot{q}}+\bm{g} -\bm{Gu}) 
\end{aligned}
\end{equation}
next, we extract and discretize the estimated forces part from \eqref{equ:observer_matrix_equation}, resulting in the following representation of the discrete force estimator:
\begin{equation}
\label{equ:simple_DOB}
\begin{aligned}
    \bm{\hat{f}_{e}}(k+1) = (\bm{I}-\frac{\delta{t}}{2m}\bm{K_I})\bm{\hat{f}_{e}}(k)+ \\\frac{\delta{t}}{2m}\bm{K_I} (m\bm{\ddot{p}}- \bm{g} +  u_f\bm{z_b})
\end{aligned}
\end{equation}
where $\bm{\hat{f}_e}$ is the estimated value of $\bm{f_e}$; $k$ is control sequence; $\delta{t}$ is sample time; $m$ is mass of UAV; $\lambda_i, i \in \{1,2,3\}$  are the i-th diagonal elements of $\bm{K_I}$; $\bm{z_b}$ is the third column of $\bm{ R(\eta) }$.

The estimated force $\bm{\hat{f_e}}$ is then transformed into the intermediate frame, denoted as $\bm{\hat{f}_{ce}} = [\hat{f}_{cex},\hat{f}_{cey},\hat{f}_{cez}]^T$, and serves as the input for the 3-D force-air model.

\section{FORCE-AIR MODEL FITTING}
\label{sec:Force_wind_relationship}

\subsection{Data Collecting}
\label{subsec:data_preprocessing}

In the experimental setup shown in Fig.~\ref{fig:first_page_show}, hover tests were conducted with our UAV platform within a wind tunnel to acquire data on various wind and their corresponding total states $\bm{\Lambda}$. The objective was to observe the UAV's response under varying wind conditions. For this purpose, the UAV was rotated $10^\circ$ every $20$ seconds and the total states $\bm{\Lambda}$ were recorded for a specific wind speed. This rotation process was repeated until the UAV completed a full revolution from $0^\circ$ to $360^\circ$. Subsequently, we increased the wind speed by $1$ m/s and repeated the rotation process, gradually covering a wind speed range from $0 $ m/s to $8 $ m/s. The resulting data points, representing the DOB forces and wind speeds, are shown as blue dots in Fig. \ref{fig:force_fitting_polar} (a).
After data clean \cite{bansal2022nearestNeighbordataclean}, the collected data was specifically used to develop the horizontal force-air model.
To gather data for the vertical force-air model, flight tests were conducted, during which the vehicle executed maneuvers in both upward and downward directions at various speeds. Throughout these flight tests, we recorded the corresponding total states $\bm{\Lambda}$.

\subsection{Data Fitting}
\label{subsec:data_fitting}

\begin{figure}[t]
	\vspace{-0.0cm}
	\centering
	\includegraphics[width=1.0\columnwidth]{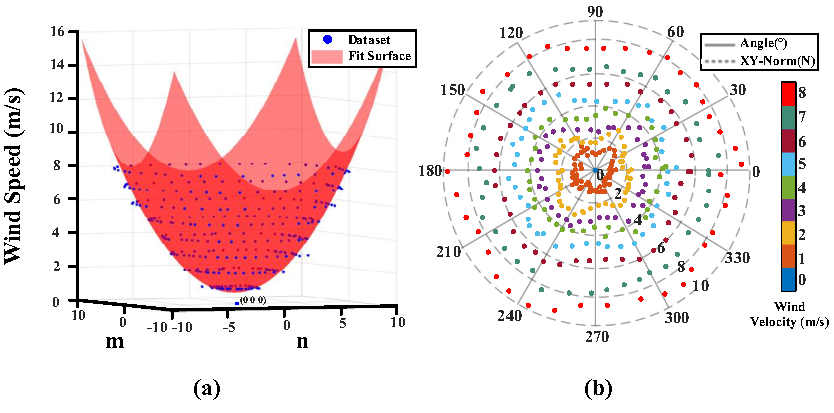}
	\caption{The horizontal force-air model. 
		\label{fig:force_fitting_polar}}
	\vspace{-0.65cm}
\end{figure}

The conventional modeling for wind speed and direction is complex. It considers the wind direction as equivalent to the external force angle on the horizontal plane, while the wind speed is a second-order function of the force norm \cite{wang2021windspeedmodelaerodynamic}. However, as discussed in Sec.\ref{subsec:data_preprocessing}, we can simplify the modeling approach. Thus, we can compute the two components as follows:
\begin{equation}
\begin{aligned}
\label{equ:fitting_model_input}
    &\vartheta_r = \arctan2(\hat{f}_{cey},\hat{f}_{cex})\\
    &f_{h}=\sqrt{\hat{f}_{cex}^2+\hat{f}_{cey}^2 }
\end{aligned}
\end{equation}
where $f_h$ represents the norm of the horizontal external force, and $\vartheta_r$ is the force angle, indicating the angle from $x_c$ to the force direction. Actually, the wind vector points towards the center of the UAV, resulting in a $180^\circ$ difference between the force direction and the wind direction.

Fig.~\ref{fig:force_fitting_polar} (a) illustrates the relations between the force angle $\vartheta_r$, the horizontal force $f_{h}$, and the horizontal wind speed $V_{wh}$ in a coplanar plane.
The blue points contain information about the correlation between the estimated external force and wind speed. The inputs $m = f_{h}\cos{\vartheta_r}$, $n = f_{h}\sin{\vartheta_r}$ correspond to the wind speed output. The force-air model is obtained by minimizing the sum of squared errors between the observed data points and their corresponding predicted values on the fitted surface, which is depicted as the red surface.
Similarly, we apply the vertical force-air model, assuming that the tilt angle of the vehicle has no impact on the vertical direction of the wind. Therefore, for the vertical force-air model, we can directly map the wind speed based on the vertical force $f_{cez}$. By combining these two models, a comprehensive 3D force-air model can be established, including all directions and utilizing the variable $f_{ce}$. The horizontal force-air surface is parameterized as follows:
\begin{equation}
\label{equ:parametrization_horizontal}
\begin{aligned}
    &V_{wh} = 0.90 + 0.06m + 0.16n + 0.09m^2 + 0.07n^2\\
\end{aligned}
\end{equation}
Fig.~\ref{fig:force_fitting_polar} (b) provides a visual representation of the dataset using polar coordinates. The polar diameter (dashed line) corresponds to $f_{h}$, the polar angle (solid line) represents $\vartheta_r$, and the depth (shown on the right color bar) signifies the magnitude of the wind speed. Due to the irregular shape of the windward side of the vehicle, the external force varies with changes in the angle. By examining the known $f_{h}$ and $\vartheta_r$ of the points positioned in each layer, we can compute the magnitude of the wind speed.

During UAV flight, the predicted vector obtained from the force-air model represents the relative air vector $\bm{A_{rc}}$ in the intermediate frame. To express $\bm{A_{rc}}$ in the inertial frame, a rotation transformation is applied:
\begin{equation}
\label{equ:fitting_final}
\begin{aligned}
    &\bm{A_{r}} = \bm{R_c}\bm{A_{rc}}\\
\end{aligned}
\end{equation}
where $\bm{R_c}$ is the rotation matrix from intermediate frame to inertial frame; $\bm{A_{r}} = [A_{rx},A_{ry},A_{rz}]^T$. As outlined in Sec.~\ref{sec:principle_of_wind_est}, the proposed dynamic wind estimation is based on the wind triangle. The environmental wind vector is determined by subtracting the UAV's ground vector from the relative air vector:
\begin{equation}
\begin{aligned}
\label{equ:dynamic_fitting_equ_1}
    \bm{A_w}  = \bm{A_r} - \bm{A_g}
\end{aligned}
\end{equation}
where $\bm{A_w}=[A_{wx},A_{wy},A_{wz}]^T$ denotes the desired wind vector; $\bm{A_g}=[A_{gx},A_{gy},A_{gz}]^T$ is the UAV's ground vector obtained from GPS. All quantities are described in the inertial frame of reference. Based on the previous explanation, we can derive the wind speed and wind direction from:
\eqref{equ:dynamic_fitting_equ}:
\begin{equation}
\begin{aligned}
\label{equ:dynamic_fitting_equ}
    &\vartheta_w = \arctan2(A_{wy},A_{wx})\\
    &V_{wh} = \sqrt{A_{wx}^2+A_{wy}^2}\\
    &V_{wv} = A_{wz}\\
\end{aligned}
\end{equation}

Ultimately, while the UAV is in flight, we can estimate the ambient wind direction $\vartheta_w$, as well as wind speed $\bm{V_w} = [V_{wh}, V_{wv}]^T$ in the horizontal and vertical directions, respectively.


\subsection{Dynamic Filter}
\label{subsec:dynamic_filter}

The external force obtained by DOB encompasses high-frequency noise stemming from sensor drift and differential noise. Building upon the results of \cite{zLyu2020dobVtol}, we identified frequency response is almost equal to $1$ over low frequencies up to $0.5$ Hz,  indicating that the filter is capable of estimating wind at this frequency. To attenuate the high frequency noise, the $0.5$ Hz disturbance bandwidth is adequate to cover the wind considered in this study. 
During periods of low wind speeds, increasing the filter gain is essential to mitigate noise. Conversely, in regions of higher wind speeds, reducing the filter gain becomes necessary to minimize phase delay and amplitude decline. To address these challenges, we incorporate gain scheduling to dynamically adjust the filter gain based on the wind speed.
\begin{figure}[t]
	\centering
	\includegraphics[width=0.95\columnwidth]{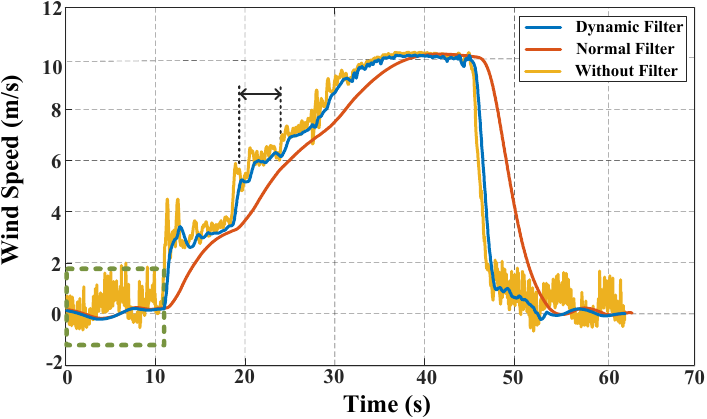}
	\caption{Estimated wind speed comparison of three filters across a wind speed range of $0-10$ m/s.
		\label{fig:dynamic_filter}}
	\vspace{-0.95cm}
\end{figure}
Fig. \ref{fig:dynamic_filter} showcases the efficacy of wind speed estimation using three filters: the dynamic filter (employing frequency gain scheduling), the normal filter (with an invariant frequency), and without a filter. The green box illustrates the dynamic filter's ability to suppress high-frequency noise compared to the results obtained without a filter. As the wind speed gradually increases to $10$ m/s, the dynamic filter reduces the lag time by $2.2$ seconds compared to the normal filter, as indicated by the black arrow. The dynamic filter outperforms both baseline approaches in noise suppression and hysteresis reduction.

\section{EXPERIMENT AND RESULTS}
\label{sec:Experiment}

\subsection{Platform Introduction}
\label{subsec:Platform Design}

As shown in Fig. \ref{fig:m300_combine_coord}, we utilized a DJI Matrice 300 UAV for our experiments. To improve the accuracy of wind estimation, a wind barrel was incorporated directly under the UAV. This addition increases the wind resistance experienced by the UAV, allowing for more precise wind estimation results. The maximum flight speed of the modified UAV is $15$ m/s. To ensure safety during wind tunnel flights, a margin space of $5$ m/s has been reserved, allowing for a maximum airflow speed measurement capability of $10$ m/s.

\subsection{Force Estimation Verification}

We assess the effectiveness of the DOB in accurately estimating unidirectional pulling forces. To conduct the evaluation, we used a tensiometer with a maximum force capacity of $3$ kg and applied tensile forces along the $x_i$, $y_i$, and $z_i$ axes individually. These constant forces were simultaneously exerted for a duration of $10$ seconds. 
The estimated results of the external force are presented in Table \ref{tab:dob_test}.
The mean errors along the X and Y axes are within $1$ N, corresponding to relative errors of less than $5.5\%$ and $7.3\%$, respectively. The estimation error along the Z axes is below $2$ N (UAV weight: $8.0$ kg), corresponding to a relative error of $14.7\%$.
The variances of all the estimated values are under $0.2$ N$^2$. This confirms that our estimated external forces of the XYZ axes stayed close to the actual external forces without drifting.

\begin{figure}[t]
	\vspace{0.0cm}
	\centering
	\includegraphics[width=0.9\columnwidth]{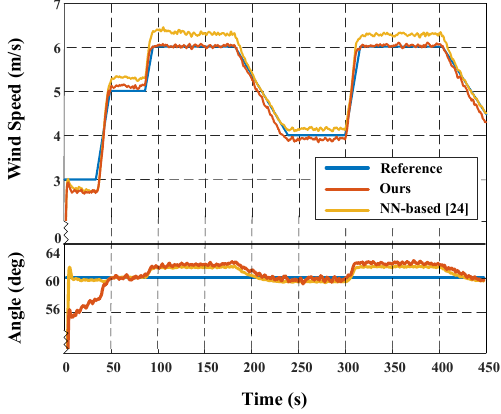}
	\caption{ Comparison of the results of the two methods for estimating $V_{wh}$ and $\vartheta_w$.
    }
		\label{fig:wind_tunnel_experiment_results}
	\vspace{-0.3cm}
\end{figure}

\begin{table}[ht]
    \vspace{0.4cm}
    \centering
    \caption{ The mean error and variance of the estimated external force along the XYZ axes in the inertial frame.}
    \label{tab:dob_test}
    \resizebox{0.78\linewidth}{!}{
        \begin{tabular}{@{}ccccc@{}}
            \toprule
            & $\hat{f}_{cex}$ & $\hat{f}_{cey}$ & $\hat{f}_{cez}$  \\ \midrule
            Reference Force (N) & 4.89 & 9.78 & 12.71 \\
            Mean Error (N) & -0.27 & -0.71 & -1.87 \\
            Variance (N$^2$) & 0.09 & 0.12 & 0.19 \\
            \bottomrule
        \end{tabular}
    }
    \vspace{-0.7cm}
\end{table}
Errors in external force estimation can be attributed to inaccuracies in the inputs specified by \eqref{equ:simple_DOB}. A notable source originates from errors in $u_f$ introduced by 'Revolution-Thrust' curve. If there are appreciable errors exist between estimated and actual external forces, this does not degrade the accuracy of the wind estimation. 
As discussed in Sec. \ref{subsec:data_fitting}, wind speed is modeled solely from the estimated force. From the variances in Table \ref{tab:dob_test}, we can see that the estimated forces on all three axes show no obvious drift. These stable errors ensure a one-to-one correspondence between the estimated forces and the actual forces. The wind speed associated with the actual external force aligns with the wind speed determined by DOB. Consequently, the estimated external force does not influence the accuracy of the wind estimation, which isolates the inaccuracy between $\bm{{f}_e}$ and $\bm{\hat{f}_e}$ from the wind estimation error.

\subsection{Wind Tunnel Verification}
\label{subsec:wind_tunnel_verification}
To validate the effectiveness of our platform, we conducted experiments in a wind tunnel. The UAV maintained a fixed wind angle (in $60 ^\circ$) while varying the wind speed over time, as shown by the blue line in Fig. \ref{fig:wind_tunnel_experiment_results}. We implemented the NN-based approach \cite{icra2023rexiansensorDeepLearning} for wind estimation on our platform, enabling a comparative analysis.
\begin{figure}[t]
	\vspace{0.0cm}
	\centering
	\includegraphics[width=1.0\columnwidth]{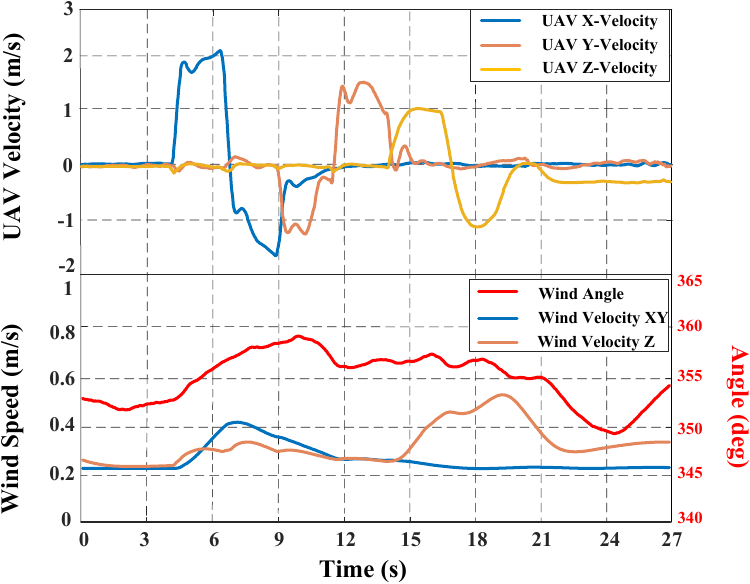}
	\caption{ Wind estimation results for UAV dynamic flight in indoor windless environment.}
		\label{fig:field_indoors_velocity_test}
	\vspace{-0.85cm}
\end{figure}
The results are presented in Fig. \ref{fig:wind_tunnel_experiment_results}. In the high wind speed region ($4-6$ m/s), our methodology achieves a mean speed/angle error of $0.09$ m/s and $2.8^\circ$, whereas the NN-based approach incurs the error of $0.33$ m/s and $2.3^\circ$, respectively. In the low wind speed region ($3$ m/s), our method exhibits a mean speed error of $0.27$ m/s and an angle error of $3.7^\circ$ versus $0.28$ m/s and $0.6^\circ$ for the NN-based approach. 
The results demonstrate that the DOB-based method outperforms the NN-based approach across both high and low wind speeds regions. The estimated angle values are also within an acceptable range.
The data extracted from the entire experiment are summarized in Table \ref{tab:estimate_wind_velocity_in_wind_tunnel}. It provides a summary account of the results, with minimum values in each row highlighted in boldface.
The analysis of the estimated velocity error indicates that the DOB-based method achieved a mean error of $0.11$ m/s and maximum error of $0.21$ m/s, while the NN-based yielded mean and maximum errors of $0.28$ m/s and $0.43$ m/s, respectively. Compared to the NN-based, the estimated mean and maximum velocity errors are improved by $60\%$ and $51\%$, respectively, using our presented DOB-based. Our method produced a mean angle error of $2.8^\circ$ and a maximum error of $5.3^\circ$, compared to $2.3^\circ$ and $2.5^\circ$ of the NN-based, respectively. 
Errors here may arise from wind flows through the vehicle, causing the attitude controller to durably adjust the attitude by varying the exposed windward area. This induces fluctuations in DOB's estimation of external forces, resulting in estimated angles and speeds that oscillate around the reference values.
\begin{table}[ht]
	\vspace{0.2cm}
	\centering
	\caption{The overall mean and maximum errors of the estimated velocity and angle for the two methods are derived from Fig. \ref{fig:wind_tunnel_experiment_results}. }
	\label{tab:estimate_wind_velocity_in_wind_tunnel}
	\resizebox{0.8\linewidth}{!}{
		\begin{tabular}{@{}ccccc@{}}
			\toprule
			& Ours & NN-based \\ \midrule
			Mean Speed Error (m/s) &\textbf{{0.11}} & 0.28 \\
			Max Speed Error (m/s) & \textbf{{0.21}} & 0.43  \\
			Mean Angle Error (deg) & 2.8  &\textbf{{2.3}} \\
			Max Angle Error (deg) & 5.3 & \textbf{{2.5}} \\
			\bottomrule
		\end{tabular}
	}
	\vspace{-0.2cm}
\end{table} 

The DOB-based wind estimation exhibits low sensitivity to errors in the estimated force, rendering the estimation results robust against such noise. In addition, it shows high adaptability to various UAV configurations as the parameters are straightforward to adjust for different aircraft designs.

\vspace{0.1cm}
\subsection{Field Test Verification}
\subsubsection{Indoor flight test}

We validated the accuracy of the wind estimation during the dynamic flight of the vehicle in an indoor no-wind environment. As shown at the top of Fig. \ref{fig:field_indoors_velocity_test}, the vehicle hovered for the first $4$ seconds. It then began dynamic flight, traversing to and fro, left and right, as well as up and down. While hovering, the mean $\hat{V}_{wh}$ is $0.21$ m/s, the mean $\hat{V}_{wv}$ is $0.22$ m/s, and the mean $\hat{\vartheta}_{r}$ is $353^\circ$. We adopt these values as baselines. When flying horizontally, the average $\hat{V}_{wh}$ is $0.35$ m/s and the $\hat{V}_{wv}$ is $0.32$ m/s. When flying vertically, the $\hat{V}_{wh}$ coincides with the baseline, while $\hat{V}_{wv}$ is $0.40$ m/s. Throughout, $\hat{\vartheta}_{r}$ fluctuates within the bounds of $348^\circ$ to $357^\circ$ with an average of $353^\circ$. From these data, we can see the dynamic flight introduced the mean errors of about $0.14$ m/s and $0.18$ m/s in the horizontal and vertical directions, respectively. The mean error in angle direction remained unchanged, but the distribution of estimated values become more spread out.
Overall, while the force-air model provides a robust estimate for the majority of high wind speed conditions, residual errors exist in low wind speed regimes owing to modeling restrictions. During the dynamic flight, the non-zero advance ratio of the propellers makes the predicted thrust $\hat{u_f}$ from the "Revolution-Thrust" curve higher than the actual values, thus augmenting the wind velocity. However, the angle is ascertained by the ratio of external forces, so it remained stable within a certain range.

\begin{figure}[t]
	\vspace{0.1cm}
	\centering
	\includegraphics[width=1.0\columnwidth]{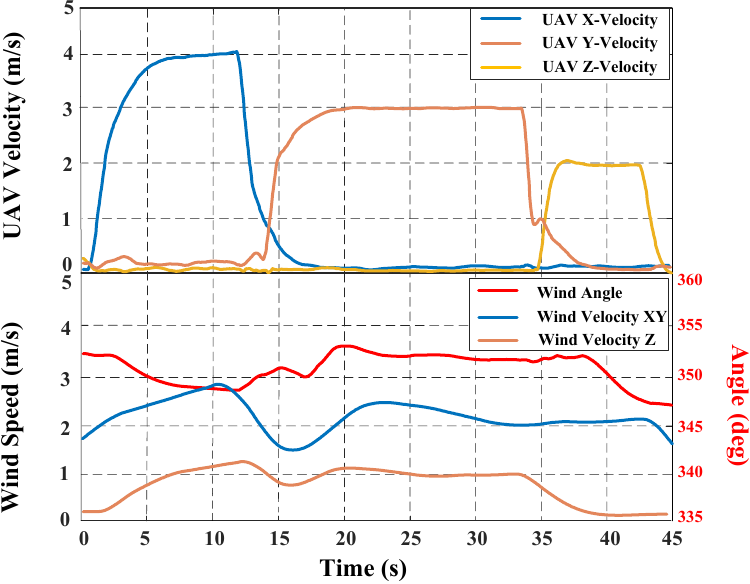}
	\caption{Wind estimation results for UAV dynamic flight in outdoors windy environment.}
		\label{fig:field_outdoors_velocity_test}
	\vspace{-1.0cm}
\end{figure}
\subsubsection{Outdoor flight test}

We validate the capability of our platform to estimate wind under outdoor dynamic flight conditions. The UAV velocities and estimated results are presented in Fig. \ref{fig:field_outdoors_velocity_test}. Despite potential errors in the anemometer measurements, we observed a wind speed range of $2-3$ m/s, a northward wind direction ($360^\circ$ in the inertial frame), and a slight vertical wind speed. During dynamic flight maneuvers, the mean $\hat{V}_{wh}$ is $2.2$ m/s, the average $\hat{V}_{wv}$ is $1.0$ m/s, and the average $\hat{\vartheta}_{r}$ is $352^\circ$. The field conditions show a time-varying wind pattern characterized by smooth transition. The results show that our estimated values exhibited smooth attributes consistent with expectations. 

\section{CONCLUSION AND FUTURE WORK}
\label{sec:conclusion}

Achieving high-precision wind estimation using only onboard sensors without specialized equipment remains an active challenge in robotics. This paper addresses this challenge by proposing a DOB-based methodology to enable UAVs to accurately determine ambient wind vectors during flight.
We designed a bespoke wind barrel to augment the accuracy of wind estimation. Our platform partitions the force estimator module and the force-air module into front-end and back-end segments individually. This architecture allows for separate optimization of the algorithms in each module. Sec. \ref{subsec:dynamic_filter} demonstrates that our approach can estimate wind speeds ranging from $10$ m/s with real-time performance. Wind tunnel tests confirm that our method achieves wind speed and angle estimation accuracies of $0.11$ m/s and $2.8^\circ$ respectively, and the estimated wind speed has $60\%$ improvements over the neural network-based method. Based on the wind triangle synthesis, our platform is qualified to estimate the 3-D ambient wind (in horizontal and vertical direction) during dynamic flight. Experiments under windless and windy conditions shows that our platform introduce only a bias of $0.2$ m/s during dynamic flight, and is able to capture the characteristics of the real world wind.
However, challenges remain in accurately modeling conditions at low wind speeds (below $1$ m/s) where the physical model is compromised. Additionally, the absence of well-defined vertical wind components complicates 3-D estimation during dynamic flight. Future work will focus on exploring torque-air model and refining model fitting approaches to enhance low-speed estimation accuracy and dynamic wind vector tracking.

\bibliography{RAL2018_reference} 

\begin{thebibliography}{10}
\providecommand{\url}[1]{#1}
\csname url@rmstyle\endcsname
\providecommand{\newblock}{\relax}
\providecommand{\bibinfo}[2]{#2}
\providecommand\BIBentrySTDinterwordspacing{\spaceskip=0pt\relax}
\providecommand\BIBentryALTinterwordstretchfactor{4}
\providecommand\BIBentryALTinterwordspacing{\spaceskip=\fontdimen2\font plus
\BIBentryALTinterwordstretchfactor\fontdimen3\font minus \fontdimen4\font\relax}
\providecommand\BIBforeignlanguage[2]{{%
\expandafter\ifx\csname l@#1\endcsname\relax
\typeout{** WARNING: IEEEtran.bst: No hyphenation pattern has been}%
\typeout{** loaded for the language `#1'. Using the pattern for}%
\typeout{** the default language instead.}%
\else
\language=\csname l@#1\endcsname
\fi
#2}}

\bibitem{patrikar2020windIrosPFGFRCFD}
J.~Patrikar, B.~G. Moon, and S.~Scherer, ``Wind and the city: Utilizing uav-based in-situ measurements for estimating urban wind fields,'' in \emph{2020 IEEE/RSJ International Conference on Intelligent Robots and Systems (IROS)}.\hskip 1em plus 0.5em minus 0.4em\relax IEEE, 2020, pp. 1254--1260.

\bibitem{cho2011windHybridSensorwind}
A.~Cho, J.~Kim, S.~Lee, and C.~Kee, ``Wind estimation and airspeed calibration using a uav with a single-antenna gps receiver and pitot tube,'' \emph{IEEE transactions on aerospace and electronic systems}, vol.~47, no.~1, pp. 109--117, 2011.

\bibitem{lyu2018dobHinfinity}
X.~Lyu, J.~Zhou, H.~Gu, Z.~Li, S.~Shen, and F.~Zhang, ``Disturbance observer based hovering control of quadrotor tail-sitter vtol uavs using $\mathrm{H}_\infty$ synthesis,'' \emph{IEEE Robotics and Automation Letters}, vol.~3, no.~4, pp. 2910--2917, 2018.

\bibitem{chen2000nonlinearChenwenhua}
W.-H. Chen, D.~J. Ballance, P.~J. Gawthrop, and J.~O'Reilly, ``A nonlinear disturbance observer for robotic manipulators,'' \emph{IEEE Transactions on industrial Electronics}, vol.~47, no.~4, pp. 932--938, 2000.

\bibitem{NEUMANN2015tiltearliest}
\BIBentryALTinterwordspacing
P.~P. Neumann and M.~Bartholmai, ``Real-time wind estimation on a micro unmanned aerial vehicle using its inertial measurement unit,'' \emph{Sensors and Actuators A: Physical}, vol. 235, pp. 300--310, 2015. [Online]. Available: \url{https://www.sciencedirect.com/science/article/pii/S0924424715301539}
\BIBentrySTDinterwordspacing

\bibitem{baker2014lidar3DwindMeasurement}
W.~E. Baker, R.~Atlas, C.~Cardinali, A.~Clement, G.~D. Emmitt, B.~M. Gentry, R.~M. Hardesty, E.~K{\"a}ll{\'e}n, M.~J. Kavaya, R.~Langland, \emph{et~al.}, ``Lidar-measured wind profiles: The missing link in the global observing system,'' \emph{Bulletin of the American Meteorological Society}, vol.~95, no.~4, pp. 543--564, 2014.

\bibitem{cho2011windfixedwingPitotTube}
A.~Cho, J.~Kim, S.~Lee, and C.~Kee, ``Wind estimation and airspeed calibration using a uav with a single-antenna gps receiver and pitot tube,'' \emph{IEEE transactions on aerospace and electronic systems}, vol.~47, no.~1, pp. 109--117, 2011.

\bibitem{makaveev2023vtolsensorwind}
M.~Makaveev, M.~Snellen, and E.~J. Smeur, ``Microphones as airspeed sensors for unmanned aerial vehicles,'' \emph{Sensors}, vol.~23, no.~5, p. 2463, 2023.

\bibitem{palomaki2017tiltwindHighReference}
R.~T. Palomaki, N.~T. Rose, M.~van~den Bossche, T.~J. Sherman, and S.~F. De~Wekker, ``Wind estimation in the lower atmosphere using multirotor aircraft,'' \emph{Journal of Atmospheric and Oceanic Technology}, vol.~34, no.~5, pp. 1183--1191, 2017.

\bibitem{meier2022windtiltnewAtmosphere}
K.~Meier, R.~Hann, J.~Skaloud, and A.~Garreau, ``Wind estimation with multirotor uavs,'' \emph{Atmosphere}, vol.~13, no.~4, p. 551, 2022.

\bibitem{simma2020iris}
M.~Simma, H.~Mj{\o}en, and T.~Bostr{\"o}m, ``Measuring wind speed using the internal stabilization system of a quadrotor drone,'' \emph{Drones}, vol.~4, no.~2, p.~23, 2020.

\bibitem{hattenberger2022KFestimating}
G.~Hattenberger, M.~Bronz, and J.-P. Condomines, ``Estimating wind using a quadrotor,'' \emph{International Journal of Micro Air Vehicles}, vol.~14, p. 17568293211070824, 2022.

\bibitem{gonzalez2020SystemIdentify}
J.~Gonz{\'a}lez-Rocha, S.~F. De~Wekker, S.~D. Ross, and C.~A. Woolsey, ``Wind profiling in the lower atmosphere from wind-induced perturbations to multirotor uas,'' \emph{Sensors}, vol.~20, no.~5, p. 1341, 2020.

\bibitem{loubimov2020SystemIdentify2}
G.~Loubimov, M.~P. Kinzel, and S.~Bhattacharya, ``Measuring atmospheric boundary layer profiles using uav control data,'' in \emph{AIAA Scitech 2020 Forum}, 2020, p. 1978.

\bibitem{zimmerman2022windMachineLearning1}
S.~Zimmerman, M.~Yeremi, R.~Nagamune, and S.~Rogak, ``Wind estimation by multirotor dynamic state measurement and machine learning models,'' \emph{Measurement}, vol. 198, p. 111331, 2022.

\bibitem{crowe2020earlyMachineLearning}
D.~Crowe, R.~Pamula, H.~Y. Cheung, and S.~F. De~Wekker, ``Two supervised machine learning approaches for wind velocity estimation using multi-rotor copter attitude measurements,'' \emph{Sensors}, vol.~20, no.~19, p. 5638, 2020.

\bibitem{di2022ladcIcraGuo3Disturbance}
J.~Di, S.~Chen, H.~Yan, X.~Wang, H.~Zhang, H.~Ji, and T.~Jin, ``Ladc: Learning-based anti-disturbance control for washing drone,'' in \emph{2022 International Conference on Robotics and Automation (ICRA)}.\hskip 1em plus 0.5em minus 0.4em\relax IEEE, 2022, pp. 5886--5892.

\bibitem{emran2018dynamicmodellagelangri}
B.~J. Emran and H.~Najjaran, ``A review of quadrotor: An underactuated mechanical system,'' \emph{Annual Reviews in Control}, vol.~46, pp. 165--180, 2018.

\bibitem{deters2014reynoldsLeinuoshu2}
R.~W. Deters, G.~K. Ananda~Krishnan, and M.~S. Selig, ``Reynolds number effects on the performance of small-scale propellers,'' in \emph{32nd AIAA applied aerodynamics conference}, 2014, p. 2151.

\bibitem{yuksel2014nonlinearFourwrench}
B.~Y{\"u}ksel, C.~Secchi, H.~H. B{\"u}lthoff, and A.~Franchi, ``A nonlinear force observer for quadrotors and application to physical interactive tasks,'' in \emph{2014 IEEE/ASME international conference on advanced intelligent mechatronics}.\hskip 1em plus 0.5em minus 0.4em\relax IEEE, 2014, pp. 433--440.

\bibitem{bansal2022nearestNeighbordataclean}
M.~Bansal, A.~Goyal, and A.~Choudhary, ``A comparative analysis of k-nearest neighbor, genetic, support vector machine, decision tree, and long short term memory algorithms in machine learning,'' \emph{Decision Analytics Journal}, vol.~3, p. 100071, 2022.

\bibitem{wang2021windspeedmodelaerodynamic}
Y.~Wang, R.~Zou, F.~Liu, L.~Zhang, and Q.~Liu, ``A review of wind speed and wind power forecasting with deep neural networks,'' \emph{Applied Energy}, vol. 304, p. 117766, 2021.

\bibitem{zLyu2020dobVtol}
M.~Zheng, X.~Lyu, X.~Liang, and F.~Zhang, ``A generalized design method for learning-based disturbance observer,'' \emph{IEEE/ASME Transactions on Mechatronics}, vol.~26, no.~1, pp. 45--54, 2020.

\bibitem{icra2023rexiansensorDeepLearning}
N.~Simon, A.~Z. Ren, A.~Piqué, D.~Snyder, D.~Barretto, M.~Hultmark, and A.~Majumdar, ``Flowdrone: Wind estimation and gust rejection on uavs using fast-response hot-wire flow sensors,'' in \emph{2023 IEEE International Conference on Robotics and Automation (ICRA)}, 2023, pp. 5393--5399.

\end{thebibliography}
\end{document}